\documentclass{article}

\usepackage{PRIMEarxiv}

\usepackage[utf8]{inputenc} % allow utf-8 input
\usepackage[T1]{fontenc}    % use 8-bit T1 fonts
\usepackage{hyperref}       % hyperlinks
\usepackage{url}            % simple URL typesetting
\usepackage{booktabs}       % professional-quality tables
\usepackage{nicefrac}       % compact symbols for 1/2, etc.
\usepackage{microtype}      % microtypography
\usepackage{lipsum}
\usepackage{fancyhdr}       % header
\usepackage{graphicx}       % graphics
\usepackage{adjustbox}
\usepackage{amsthm}
\usepackage{amssymb}
\usepackage{amsfonts}       % blackboard math symbols
\usepackage{subfigure}
\usepackage{amsmath}
\usepackage{xcolor}
\graphicspath{{media/}}     % organize your images and other figures under media/ folder

\title{GhostShiftAddNet: More Features from Energy-Efficient Operations
}

\author{
  Jia Bi, Jonathon Hare, Geoff V. Merrett\\
  Electronics and Computer Science \\
  University of Southampton \\
  Southampton\\
  \texttt{\{}{J.Bi, jsh2, gvm\}@ecs.soton.ac.uk} \\
}

\begin{document}

\maketitle

\begin{abstract}
Deep convolutional neural networks (CNNs) are computationally and memory intensive. In CNNs, intensive multiplication can have resource implications that may challenge the ability for effective deployment of inference on resource-constrained edge devices. This paper proposes GhostShiftAddNet, where the motivation is to implement a hardware-efficient deep network: a multiplication-free CNN with fewer redundant features. We introduce a new bottleneck block, GhostSA, that converts all multiplications in the block to cheap operations. The bottleneck uses an appropriate number of bit-shift filters to process intrinsic feature maps, then applies a series of transformations that consist of bit-wise shifts with addition operations to generate more feature maps that fully learn to capture information underlying intrinsic features. We schedule the number of bit-shift and addition operations for different hardware platforms. We conduct extensive experiments and ablation studies with desktop and embedded (Jetson Nano) devices for implementation and measurements. We demonstrate the proposed GhostSA block can replace bottleneck blocks in the backbone of state-of-the-art networks architectures and gives improved performance on image classification benchmarks. Further, our GhostShiftAddNet can achieve higher classification accuracy with fewer FLOPs and parameters (reduced by up to $3\times$) than GhostNet. When compared to GhostNet, inference latency on the Jetson Nano is improved by 1.3$\times$ and 2$\times$ on the GPU and CPU respectively. Code is available open-source on \url{https://github.com/JIABI/GhostShiftAddNet}.
\end{abstract}

%-------------------------------------------------------------------------
\section{Introduction}
Deep Convolutional Neural Networks (CNNs) have become more accurate and faster for image classification applications with large image datasets. However, traditional CNN networks with many parameters and floating point operations (FLOPs) are problematic to deploy on resource-constrained hardware platforms within specific application scenarios (e.g. low latency requirements are a priority for autonomous driving systems). To address this problem, many portable networks, e.g., ShuffleNet~\cite{Zhang_2018_CVPR,Ma_2018_ECCV}, MobileNet~\cite{howard2017mobilenets,Sandler_2018_CVPR,Howard_2019_ICCV}, and CNNs based on shift or addition operation (e.g., ShiftNet~\cite{Wu2018ShiftAZ}, AdderNet~\cite{Chen_2020_CVPR}) have been proposed, requiring fewer FLOPs and parameters. However, in practice, these methods are inefficient to implement onto different hardware platforms for two reasons: the structure of convolutional components and operation selection in CNNs. 

Firstly, three important factors in achieving a hardware efficient network are a small number of FLOPs and parameters, and a low inference latency. However, common convolution components in CNNs, including Spatial Convolution (SConv), Depth Separable Convolution (DWSConv), and Shift Convolution (ShiftConv), cannot satisfy these three factors concurrently. For example, because SConv requires a considerable number of FLOPs, they are inefficient and limited to use on compute-bound hardware platforms (e.g. CPU)~\cite{Chen2019AllYN}. Compared with SConv, DWSConv and ShiftConv require fewer FLOPs and parameters but cannot effectively reduce inference latency, especially on memory-bound platforms (e.g. GPU). The main reason for this is that DWSConv and ShiftConv require more memory accesses than computation~\cite{DBLP:conf/wacv/HeLZM19,Ma_2018_ECCV,Chen2019AllYN}. As a result, a first question naturally arises: \textit{What is the best structure of the convolutional component in CNNs, which allows the network to run efficiently on CPU- and memory-bound platforms?} 

Secondly, CNNs contain many multiplications, and their high computational load prevents them from running on embedded platforms with limited power budgets. Therefore, considerable research has aimed to replace multiplication operations with ``cheap'' operations (e.g., shift and addition operations)~\cite{NEURIPS2020_1cf44d79}. Although addition operations have significantly fewer FLOPs than multiplications, we observe that CNNs based on addition operations may have a longer training process and higher inference latency than CNNs based on multiplication. This is for two possible reasons: Firstly, during the training process, the loss function of CNNs that use addition is measured by the $\ell_{1}$ distance and the magnitude of the variance of the gradient of the loss function is larger than that of multiplication. This leads to slower convergence of the network and lower accuracy~\cite{Chen_2020_CVPR}. Secondly, addition operations in floating point (FP) format may actually have higher inference latency than multiplication operations; following the IEEE754 FP standard~\cite{8766229}, according to the difference between the two exponents, the FP addition operation needs to align the two mantissas to be added. This may require multiple variable shifts before the adder. The result of mantissa addition then needs to be renormalized, also requiring many shifts to format the result correctly. Therefore, compared to multiplication, adding two mantissas may require higher gate delays and line delays.
%FP addition requires to shift the mantissa of the number for exponents match, and such normalization process is needed to repeat three times in series, but FP multiplication only requires to multiply the mantissa and add exponents, and these two steps are done in parallel. As a result, we observed that addition operations in FP format have a longer latency than FP multiplication using Intellectual Property core. For example, an addition operation always takes 11 and 12 clock Cycles in non-blocking and blocking modes, respectively, while a multiplication only takes ~8 and 9 clock Cycles. 
As a result, if aiming to use cheap operations to replace multiplications in CNNs, a second question is posed: \textit{what is the efficient structure of operations in a convolutional component?}

\begin{figure}
\begin{center}
\includegraphics[height=2.5cm]{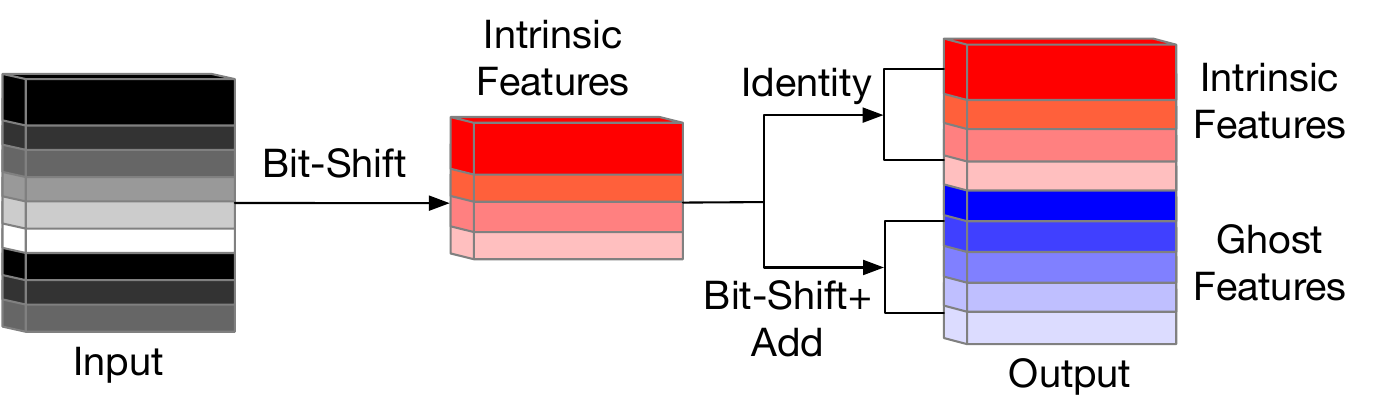}
\end{center}
\caption{An illustration of proposed GhostSA module for outputting the same number of features. SConv applies the number of bit-shift filters to generate intrinsic features from the input features. Then a modified DWSConv utilises bit-wise shift and addition operations~\cite{NEURIPS2020_1cf44d79} to generate a series of ghost features.}
\label{ghostModule}
\end{figure}

These two issues prompted us to propose a novel CNN \textit{GhostShiftAddNet} (GhostSANet), a lightweight network topology with hardware-friendly convolution operations, which can be efficiently implement on embedded CPUs and GPUs. In particular, we designed a new module called \textit{GhostShiftAdd} (GhostSA), partly inspired by GhostNet~\cite{Han_2020_CVPR} and ShiftAddNet~\cite{NEURIPS2020_1cf44d79}, which divides the convolutional output layer into two parts, as shown in red and blue in Figure~\ref{ghostModule}. In this figure, some bit-wise shift operations identify the entire features of the input layer (called ``intrinsic'' features) and convolve them into the red part. Given the intrinsic features from the red part, the blue part can be obtained by using the modified DWSConv and cheap operations to generate more ``ghost'' features based on the intrinsic features. The output feature consists of concatenating two parts, the size of which has not changed. Given the size of the output layer, a hyper-parameter $\gamma$ balances the number of intrinsic and ghost features in the output layer so that the number of ghost features can be controlled to enrich the intrinsic feature information from the input layer, and feature redundancy can also be avoided.
%As a result, GhostSA module based CNNs have three benefits, including reducing the large memory movements in convolutional networks by reducing the redundancy of feature maps; scheduling the number of shift and addition operations to balance the accuracy- and energy-latency trade-offs in the network; and use a self-adaptive variance reduced optimization algorithm (Variance Controlled Stochastic Gradients (VCSG)~\cite{Jiabi}) to decrease the variance of gradients generated by addition operations to fast achieve the high accuracy during training process.
Consequently, our contributions can be summarized as follow: 
\begin{enumerate}
    \item A novel multiplication-free module \textit{GhostShiftAdd} (GhostSA) uses a hyper-parameter $\gamma$ to control the number of bit-wise shift and addition operations in a convolutional layer. The module can be easily applied to any CNN architecture. This is the first work to concurrently reduce the FLOPs, memory requirements and inference latency of both traditional and portable CNNs on general-purpose and embedded hardware platforms. 
    \item The GhostSA module provides a modified DWSConvt. The improved DWSConvt uses bit-wise operation in the depth-wise layer to quickly parallel shift features of separable channels. Then, it uses addition operations in the point-wise layer to combine the feature information from all channels and adjust the parameters to avoid the accuracy loss through the shift operation in the depth layer. Our theoretical results show that the GhostSA module can significantly reduce FLOPs, memory requirements, and inference latency of multiplication-based spatial convolutions by $k^{2}\gamma\times$ (typical values are normally a filter kernel size $k=3$ and $\gamma>=2$).
    \item Based on the GhostSA module, we present \textit{GhostShiftAddNet} (GhostSANet), a novel hardware-efficient neural architecture. We demonstrate the effectiveness and efficiency of GhostSANet on visual datasets, comparing with SOTA benchmarks including both traditional CNNs (such as VGG, ResNets) and portable CNN (such as ShuffleNet, MobileNet, GhostNet). Our experimental results show that the GhostSA module reduces the number of parameters and FLOPs (reduced by ~$3\times$) in the SOTA backbone network, reduces training time and inference latency (reduced GPU latency by ~$1.5\times$), and maintains or even improves  image classification accuracy (>1\%). In addition, the CPU and GPU latency of GhostSANet on an embedded platform (Nvidia Jetson Nano) is reduced by ~$2\times$ and ~$1.3\times$, respectively.
\end{enumerate}

\section{Related Work}\label{s2}
\paragraph{\textbf{Multiplication-less Deep Networks.}}
DWSConv is a method to reduce the FLOPs by decoupling SConv into spatial and cross-channel convolution processes, which requires fewer multiplications. Alternatively, there are many works which use other operations with fewer FLOPs to replace multiplications in CNNs. For example, 
ShiftNet~\cite{Wu2018ShiftAZ} generates feature maps by assigning one value in a $k\times k$ convolution kernel as 1, and the rest as 0, thereby using the shift operation to perform convolution. ~\cite{Chen2019AllYN} proposes to further reduce FLOPs by removing unnecessary shifts in ShiftNet. DeepShift~\cite{Elhoushi2019DeepShiftTM} uses bit-wise and sign-shifting instead of a standard shift during training, and only requires at most 5 bits to represent the weight. However, DeepShift reduces the relatively high classification accuracy of inference, especially on large-scale tasks (e.g., ImageNet datasets). To solve this problem, ShiftAddNet~\cite{NEURIPS2020_1cf44d79} maintains an accuracy similar to the original network by combining the bit-shift and addition operations proposed by AdderNet~\cite{Chen_2020_CVPR}. \cite{DBLP:conf/wacv/HeLZM19} observe that shift operations with fewer FLOPs do not always lead to a shorter inference latency, and hence three shift-based primitives are proposed to reduce GPU inference time. However, these methods only focus on traditional CNNs, and did not experimentally evaluate with portable CNNs due to the difficulty in improving their efficiency. This motivates us further to reduce FLOPs, parameters, and inference latency of all types of CNNs, without incurring accuracy loss.   
 
\paragraph{\textbf{Lightweight Network Topology Design.}}
In recent years, a series of efficient convolutional architectures have been proposed for lightweight network topologies. Aside from the convolution components already introduced (e.g., DWSConv, group convolution and channel shuffle~\cite{Zhang_2018_CVPR,Ma_2018_ECCV}), GhostNet~\cite{Han_2020_CVPR} aims to reduce input redundancy by using fewer convolutional filters to process input features, which is beneficial to reduce the FLOPs, parameters and memory movements in networks. Fewer memory movements can reduce inference latency in a network. Therefore, drawing on the concept of GhostNet, our fundamental idea in this paper is to identify cheap combinations of operations and efficient convolutional components architectures to build our new CNN network.

\section{The GhostShiftAdd Module: Less is More}\label{s3}
Assume input data $I\in\mathbb{R}^{c_{i}\times h_{i}\times w_{i}}$, where $c_{i}$ is the number of input channels, $h_{ i }$ and $w_{i}$ are the height and width of the input data respectively. The operation of a standard convolutional layer for producing features can be formulated as $O = I * f + b$, where $*$ is the convolution operation, $b$ is the bias term. $O\in\mathbb{R}^{h_{o}\times w_{o}\times c_{o}}$ is the output feature map with $c_{o}$ channels, and $f\in\mathbb{R}^{c_{i}\times k\times k\times c_{o}}$ is the convolution filter in this layer. Moreover, $h_{o}$ and $w_{o}$ are the height and width of the output data, and $k\times k$ is the kernel size.  During this convolution procedure, a standard convolutional network requires $c_{i}\cdot{h_{o}}\cdot{w_{o}}\cdot{c_{o}}\cdot{k}\cdot{k}$ FLOPs, which is often massive since the
number of output filters $c_{o}$ and the input channel number $c_{i}$ are generally very large (e.g. 256 or 512)~\cite{Han_2020_CVPR}. In addition, the large number of $c_{i}$ may increase memory movements resulting in high latency.

\subsection{Bit-wise shifts for Generating Intrinsic features}
Unlike GhostNet~\cite{Han_2020_CVPR}, which divides the output features into two equal parts $m1=m2$ in all cases, we provide a more general form that uses the ratio $\gamma$ to balance the two parts, such that $m_{1}=c_{o}/\gamma$ and $m_{2}=m_{1}(\gamma-1)$ ($\gamma>=2$). In Fig~\ref{ghostModule}, the red part represents the number of intrinsic features $m_{1}$ in output layer. The output intrinsic feature $O_{Int}\in\mathbb{R}^{(m_{1}\times h_{o}\times w_{o})}$ can be formulated as
\begin{equation}
    O_{Int}=f_{s}(I, w_{s}), \ \ \text{where} \ \ f_{s}(I,w_{s}) = \sum I^{T} * w_{s} \;,
    \label{e1}
\end{equation}
where $f_{s}(\cdot, \cdot)$ is a bit-wise filter that performs an inner product, and $w_{s}=s\cdot 2^{p}$ are weights in the bit-wise shift, where $p$ is an integer parameter, and $s\in\{-1, 0, 1\}$ is a sign flip operator. In particular, the bit-wise shift operation for multiplication (left shift) or division (right shift) depends on the parameter $p$. If $p$ is positive, the input value $I<<p$ means left-shift; if $p$ is negative, then right-shift. Since bit-wise shifts are always positive and there is a lack of a negative search space, the sign flip operation can expand the bit shift search space to provide a negative value (see~\cite{Elhoushi2019DeepShiftTM} for full details). Compared to GhostNet, the computational cost of bit-wise shift operations is much cheaper than multiplication. 

\subsection{Depth-wise Separable Bit-wise shifts with Adders for Generating Ghost features}
The blue part in Fig~\ref{ghostModule} represents the number of $m_{2}$ output features, namely ghost features. The ghost feature size is  $O_{ghost}\in\mathbb{R}^{(m_{2}\times h_ {o}\times w_{o})}$. These are generated by the intrinsic feature through the DWSConv structure, where the depth-wise layer uses bit-wise shift operations ($k=3$), and the point-wise layer uses addition operations ($k=1$). This idea was motivated by two reasons. Firstly, in the FP format, we find that the addition operation applied to the deep convolution filter is slower than using shift and multiplication, especially the large filter kernel size (for example, $k=3$ or $5$). However, when $k=1$, addition is as fast as shifting or multiplication. Secondly, using shift operations throughout DWSConv results in a relatively high loss in accuracy. This is because, compared with the standard convolution spanning the entire continuous space of the multiplication map in DWSConv, the bit-wise shift can only represent a subset of the power of 2 multiplication~\cite{NEURIPS2020_1cf44d79}. However, addition can effectively expand the shift parameter mapping space. Therefore, after the depth-wise layer based on bit-wise operations, we use the addition operation in the $1\times1$ point-wise layer to avoid high latencies and improve the accuracy of the bit-wise shift operations. Combining with Eq.~\ref{e1}, the ghost features part can be mathematically described as
\begin{equation}
    O_{ghost} = f_{a}(f_{s}(O_{Int}, s\cdot2^{p}),w_{a}), \ \ \text{where} \ \  f_{a}(f_{s}, w_{a}) = -\sum\|f_{s}-w_{a}\|_{1} \;,
\label{ghost}
\end{equation}
where $f_{a}(\cdot,\cdot)$ is the adder filter, and $w_{a}$ is the addition weights that is measured by $\ell_{1}$ distance. Finally, we concatenate the intrinsic and ghost features as output layer $Y= [O_{Int}, O_{ghost}]$.

\subsection{Optimization of GhostSA}
In an addition operation based CNN, the relatively large variance of the gradient would increase the FLOPs during the training process. Therefore, in our case, choosing an effective optimization method becomes very important. Stochastic Gradient Descent (SGD) and Adam are widely used in many addition networks~\cite{Chen_2020_CVPR,Song2020AdderSRTE,NEURIPS2020_1cf44d79}. %Due to its random behaviour, SGD cannot reduce or even increase the variance of the gradient in the additive network. On the other hand, 
The Adam method uses an adaptive learning rate to converge faster than SGD in some cases. 
However, the adaptive learning rate computed using the exponential moving average of the squared gradient cannot guarantee convergence in certain conditions, for example, high-dimensional settings when the variance of the gradient to time is large~\cite{ReddiKK18}. Therefore, we applied the Variance Controlled Stochastic Method (VCSG)~\cite{Jiabi} to effectively reduce the variance. This offers two improvements: the update of the adaptive learning rate switches between fixed and decayed value depending on the current variance, and a hyper-parameter controls the variance of VCSG at different stages of the training process.

\subsection{Complexity of GhostSA}
This section analyzes the performance of the GhostSA module through acceleration, model compression and memory access rates. The output channels of the intrinsic part is $m_{1}$, and the kernel size is $d$ (it is usually recommended that $d=1$). The intrinsic part requires zero FLOPs to calculate the power-of-2 function in bit-wise shift operations. Since an identity mapping has $m_{1}(m_{2}-1)$ filters for ghost features, it requires $h_{o}\cdot w_{o}\cdot m_{1}\cdot m_{ 2 }$ FLOPs. Finally, the theoretical speedup ratio of upgrading spatial convolution with our GhostSA module is
\begin{equation}
\begin{aligned}
    r_{s}&=\dfrac{c_{i}\cdot c_{o}\cdot h_{o}\cdot w_{o}\cdot k\cdot k}{h_{o}\cdot w_{o}\cdot m_{1} m_{2}}= \dfrac{c_{i}\cdot c_{o}\cdot k\cdot k}{m_{1}\cdot m_{2}}= \dfrac{k\cdot k \cdot c_{i}\cdot c_{o}}{\dfrac{c_{o}}{\gamma}\cdot\dfrac{c_{o}(\gamma-1)}{\gamma}}\approx \dfrac{ \gamma k\cdot k \cdot c_{i}}{c_{o}} \approx k^{2}\gamma \;,
\end{aligned}
\label{rs}
\end{equation}
where $d\times d = 1 \times 1$, the last approximation is obtained by assuming $\gamma << c_{i}$. Moreover, intrinsic and ghost part require $log (c_{i}\cdot d\cdot d \cdot m_{1})$ and $log (m_{1}\cdot k\cdot k) + m_{1}m_{2}$ parameters, respectively. The compression ratio of GhostSA can be formulated as
\begin{equation}
    \begin{aligned}
    r_{c}&=\dfrac{c_{i}\cdot c_{o}\cdot k\cdot k}{log (c_{i}\cdot m_{1}\cdot d\cdot d) + log (m_{1}\cdot k\cdot k)+ m_{1}\cdot m_{2}}=\dfrac{c_{i}\cdot c_{o}\cdot k\cdot k}{2log(c_{i}\cdot \dfrac{c_{o}}{\gamma}\cdot k\cdot k)+ m_{1}\cdot m_{2}}\approx k^{2}\gamma \;.
\end{aligned}
\label{rc}
\end{equation}
Finally, the memory access ratio from input to output feature maps is
\begin{equation}
    \begin{aligned}
    r_{m}&=\dfrac{c_{i}\cdot c_{o}\cdot k\cdot k\cdot h_{o}\cdot w_{o}+c_{o}\cdot h_{o}\cdot w_{o}}{h_{o}\cdot w_{o}(c_{i}\cdot m_{1}+m_{1} + m_{1}\cdot k\cdot k+ m_{1}+m_{1}\cdot m_{2}+m_{2})}= \dfrac{k\cdot k\cdot \gamma\cdot c_{i}+ c_{o}}{c_{i}+ c_{o}+ k\cdot k+1}\approx k^{2}\gamma \;.
\end{aligned}
\label{rm}
\end{equation}
In the three ratios of Eq~\ref{rs}, ~\ref{rc} and ~\ref{rm}, the kernel size $k$ is a constant value. The hyper-parameter $\gamma$ can be flexibly adjusted to control the number of shift and addition operations in each convolution layer: a key factor affecting acceleration, compression, and memory access ratios. When increasing $\gamma$, the division of ghost features becomes larger, resulting in more addition operations than shift operations in total. In this case, the same trend on both CPU and GPU is observed: that the number of parameters, FLOPs and accuracy are increased. However, the latency of the model on GPU and CPU show different trends. A larger number of shift operations on the GPU will increase the inference latency, but on the CPU will decrease latency. 

\subsection{GhostSANet: Efficient CNNs}
\begin{figure}[b]
    \centering
    \includegraphics[width=0.55\textwidth]{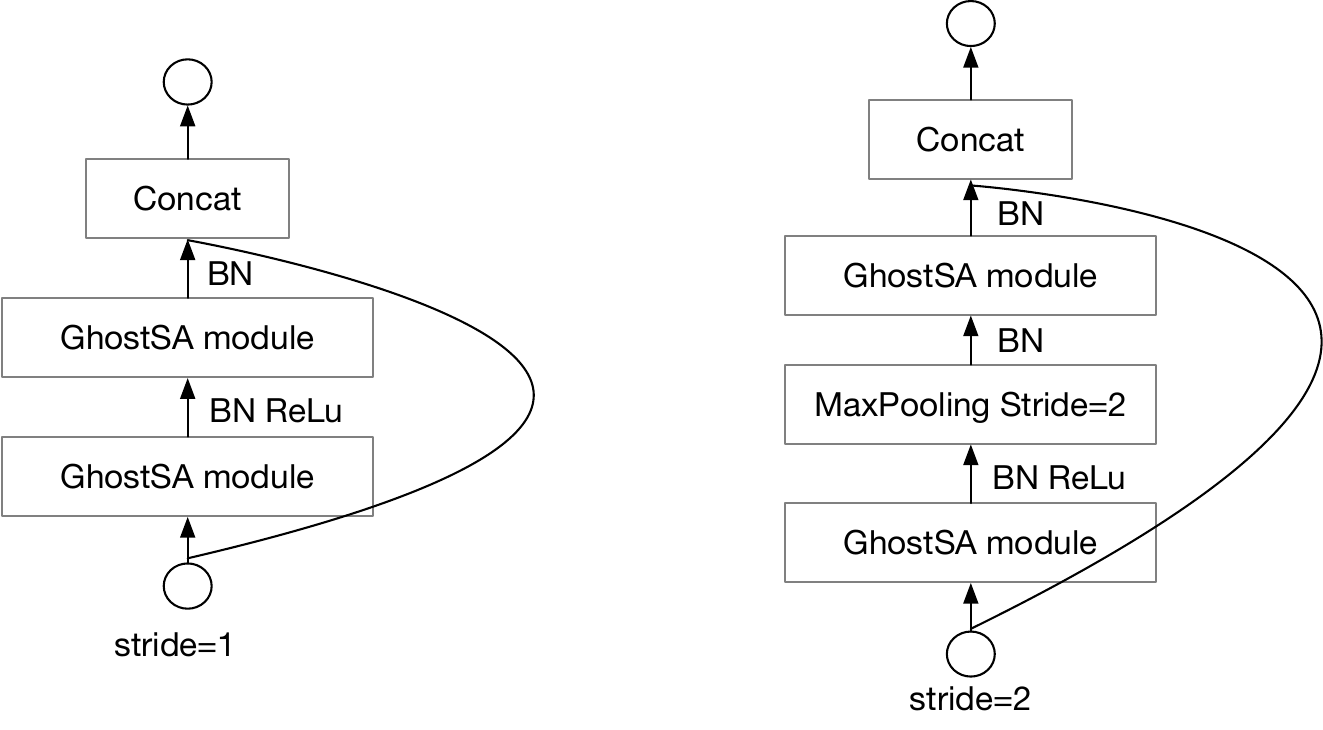}
    \caption{GhostSA bottleneck: stride=1 (left) and stride=2 (right).}
    \label{ghostsablock}
\end{figure}

\paragraph{\textbf{GhostSA Bottlenecks.}} Based on the GhostSA module, we proposed the GhostSA bottlenecks that integrate GhostSA modules and shortcuts, as shown in Fig~\ref{ghostsablock}. 
There are two versions of the GhostSA bottleneck when stride=1 and stride=2. In the case of stride=1, the GhostSA bottleneck consists of two stacked GhostSA modules. The first GhostSA module acts as an expansion layer that uses \textit{expansion ratio} to balance input and output channels. The second GhostSA module reduces the number of channels from the expansion layer to match the shortcut path. In the middle of two GhostSA modules, Batch normalization (BN)~\cite{pmlr-v37-ioffe15} is used after each layer, but only used with Relu non-linear activation layers after the first GhostSA module. For the case of stride=2, we implement the shortcut path by inserting a down-sampling layer (\textit{max-pool}) between the two GhostSA modules, which is an alternative to the depth-wise convolution structure in the Ghost bottleneck~\cite{Han_2020_CVPR}. The motivation is to simplify the network architecture and control flow, especially on edge hardware platforms.
\paragraph{\textbf{GhostSANet}}
We use the GhostSA bottleneck to build a new efficient and lightweight CNN architecture, called \textit{GhostShiftAddNet} (GhostSANet). Following from architectures of GhostNet, MobileNet v3, GhostSANet uses bit-wise shift operations in all convolutional layers, and applies GhostSA bottlenecks in the block layer, and use the ReLU non-linear activation in the first convolution layer and the GhostSA bottleneck, and apply hard-swish~\cite{Howard_2019_ICCV} in the last convolutional layer before the fully connected layer. To adapt to the application of different scales, we use the width multiplier $\alpha$ in the model, expressed as GhostSANet-$\alpha\times$, which can scale the width of the entire network.

\section{Experiments and Results Analysis}\label{s4}
%In this section, we first descirbe our experiment settings. After that, we analyse the performance of the proposed GhostSA module with different vaule of hyper-parameters $\gamma$ to verify its effectiveness by replacing the original convolutional layers of deep CNN networks by our module. Further, the new GhostSA bottleneck will be practically easily applied in the popular efficient neural networks, and be compared their performance with backbone networks. Lastly, our GhostSANet architecture built by using the new GhostSA bottleneck will be tested on the image classification benchmarks. 
\paragraph{Datasets and Architecture Settings}\label{sec4}
Our experiments are performed on image classification datasets, including the CIFAR10~\cite{Krizhevsky09} and ImageNet ILSVRC 2012 ~\cite{ILSVRC15} datasets. The CIFAR10 dataset is used to analyze the attributes of the proposed method, which consists of 60,000 32$\times$32 colour images in 10 classes, including 50,000 training images and 10,000 test images. ImageNet is a relatively large-scale image dataset containing 1.2M training images and 50K validation images in 1,000 classes. Our experiments were conducted on three hardware platforms: Intel Core i7-10700 CPU, NVIDIA GeForce RTX 2080 Ti GPU, and NVIDIA Jetson Nano (an embedded platform containing a quad-core Arm CPU, and a 128-core GPU based on the NVIDIA Maxwell architecture). For direct comparisons, we re-implement all benchmarks in our experiments following our architecture settings.

\paragraph{\textbf{Analysis of Hyper-parameters of GhostSA Module}}
In order to evaluate the behaviour of shift and addition operations on heterogeneous computing devices (such as CPU and GPU), we analyze the values of hyper-parameters $\gamma$ in the GhostSA module on popular deep CNNs (VGG-16~\cite{Simonyan15} and ResNet-20~\cite{7780459}) with the CIFAR-10 dataset. The GhostSA module replaces all convolutional layers of the reference network, and the new models are named GhostSA-VGG-16 and GhostSA-ResNet-20.

\begin{table}[htb]
\begin{center}
\resizebox{11cm}{!} {
\begin{tabular}{|c|c|c|c|c|c|c|} 
\hline
Model            & $\gamma$  & Weights (M) & FLOPs (M) & Top1-Acc. (\%) & \begin{tabular}[c]{@{}c@{}}GPU-\\Latency (ms)\end{tabular} & \begin{tabular}[c]{@{}c@{}}CPU-\\Latency (ms)\end{tabular}  \\ 
\hline\hline
ResNet-20~\cite{7780459}         & -                          & 0.27         & 41      & 92.2     & 7.5                                                       & 83                                                         \\ 
% \hline
                  & $\gamma=2$               & 0.007         & 1.12     & 87.7    & 8.7                                                       & 67                                                        \\ 
% \cline{2-7}
                  & $\gamma=3$              & 0.009         & 1.23     & 88.2     & 8.2                                                       & 81                                                        \\ 
% \cline{2-7}
GhostSA-ResNet-20 & $\gamma=4$               & 0.016         & 1.64     & 89.3     & 8.0                                                       & 99                                                         \\ 
% \cline{2-7}
                  & $\gamma=5$              & 0.021        & 2.36     & 90.3     & 7.6                                                      & 113                                                         \\ 
% \cline{2-7}
                  & $\gamma=6$               & 0.045        & 4.67     & 92.4     & 7.2                                                       & 127                                                         \\
% \hline
Ghost-ResNet-20~\cite{Han_2020_CVPR}          & -                        & 0.14          & 20      & 92.3     & 7.3                                                       & 70                                                       \\ 
% \hline
\hline
VGG-16~\cite{Simonyan15,Han_2020_CVPR}         & -                         & 15         & 313      & 93.6     & 86                                                        & 2.1                                                       \\ 
% \hline
               & $\gamma=2$                & 0.35        & 7.12     & 87.3     & 99                                                        & 1.8                                                       \\ 
% \cline{2-7}
               & $\gamma=3$                & 0.46       & 8.72     & 90.4    & 91                                                        & 2.0                                                      \\ 
% \cline{2-7}
GhostSA-VGG-16 & $\gamma=4$                & 0.57       & 12.1     & 91.6     & 87                                                        & 2.5                                                       \\ 
% \cline{2-7}
               & $\gamma=5$                & 0.92       & 17.8     & 92.1     & 83                                                        & 3.1                                                       \\ 
% \cline{2-7}
               & $\gamma=6$                & 1.8       & 35.6     & 93.5     & 78                                                        & 3.6                                                       \\
% \hline
Ghost-VGG-16~\cite{Han_2020_CVPR}        & -                         & 7.7         & 158      & 93.7     & 80                                                        & 1.9                                                       \\ 
\hline
\end{tabular}}
\end{center}
\caption{Comparison the performance of Ghost modules and GhostSA modules with different $\gamma$ for compressing ResNet20 and VGG-16 on CIFAR10. }
\label{table1}
\end{table}
In Table~\ref{table1}, we test the performance of Ghost modules applied to two backbone models named Ghost-ResNet-20 and Ghost-VGG-16, and set appropriate kernel sizes $k=3$ and $d=1$. The experimental results verify our theoretical analysis. As $\gamma$ is increased from $2$ to $6$, the partition of intrinsic features becomes smaller, resulting in the output layer containing fewer intrinsic features to represent the input features and more ghost features. More ghost features supplement information from the intrinsic part, which can increase the accuracy. In such a case, the greater number of addition operations will increase parameters and FLOPs, which places a burden on compute-bound platforms (e.g. CPU). Accordingly, the total number of bit-shift operations is relatively small, reducing memory movement on GPU. In addition, compared to the Ghost module, our module has significantly fewer FLOPs and parameters, and the accuracy loss is kept within $<1\%$, even slightly higher than the Ghost module. After many experiments and comparisons, we choose $\gamma=4$ for GPU and $\gamma=2$ for CPU in the following experiments.

\begin{figure}[htb]
\begin{center}
 \begin{tabular}{cc}
 {\includegraphics[width=0.88\textwidth]{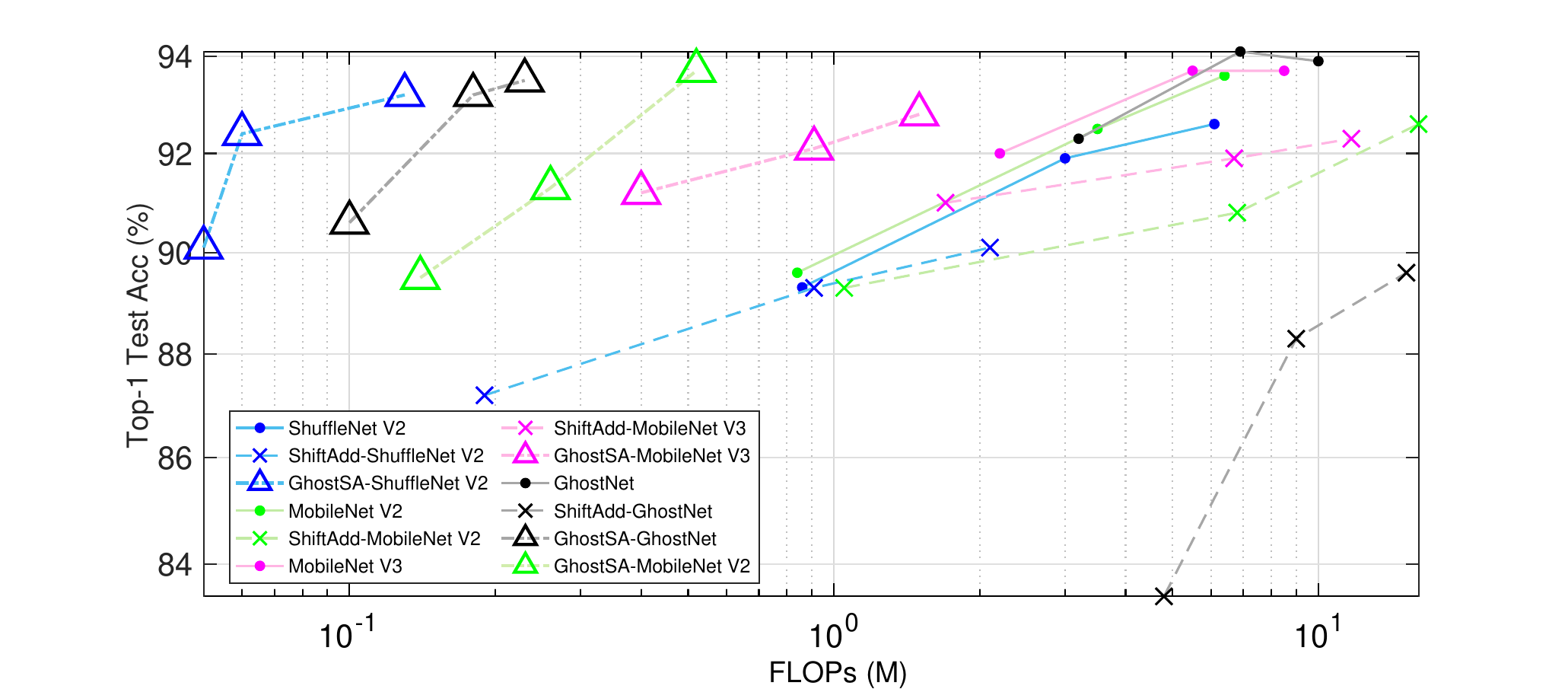}}\\
{\includegraphics[width=0.88\textwidth]{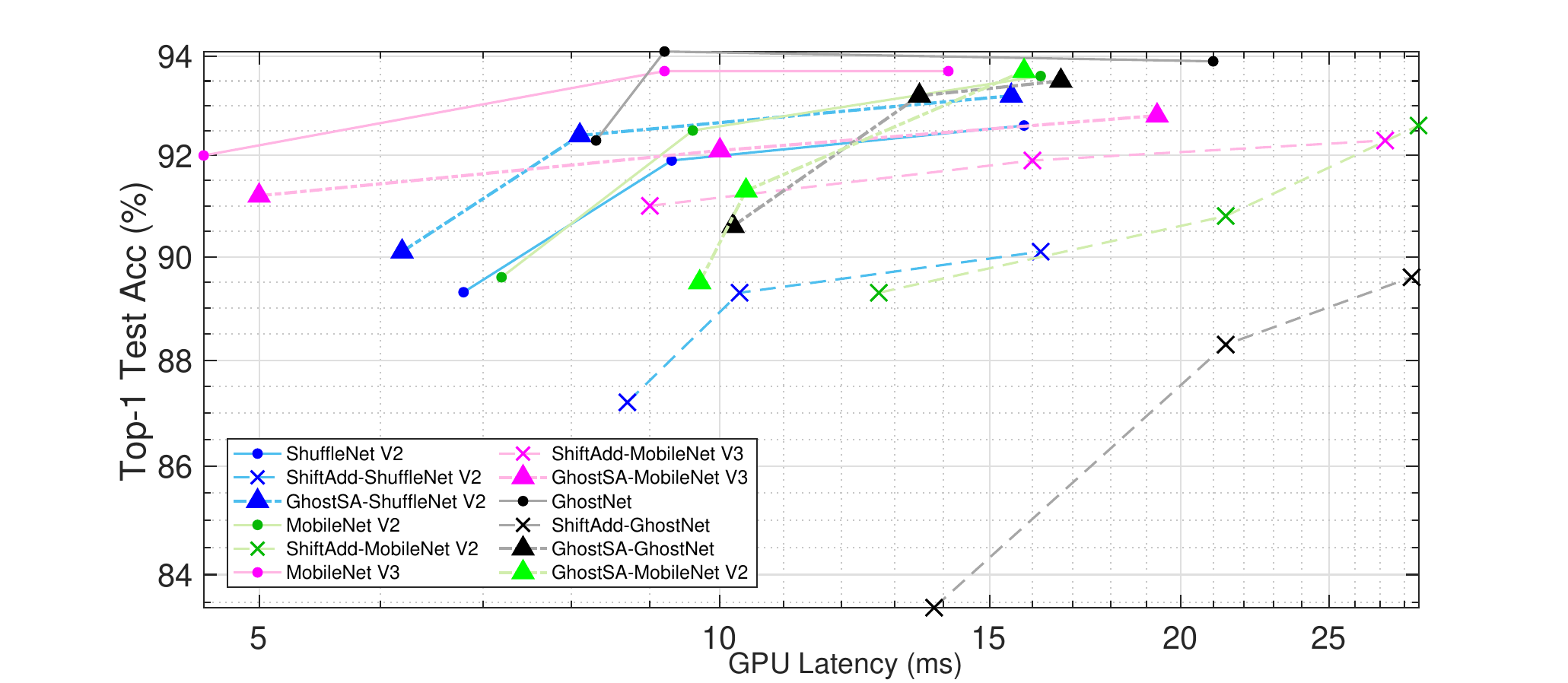}}\\
\end{tabular}
\end{center}
\caption{Comparison of FLOPs (top) and latency (bottom) of state-of-the-art lightweight networks against accuracy on the CIFAR10 dataset.}
\label{f2}
\end{figure}

\paragraph{\textbf{GhostSA Bottleneck on State-of-the-arts}}
We compare the performance of three bottlenecks including the GhostSA bottleneck, GhostNet and ShiftAddNet, by applying them in SOTA architectures, e.g., MobileNet 2~\cite{Sandler_2018_CVPR}, 3~\cite{Howard_2019_ICCV}, ShuffleNet 2~\cite{Ma_2018_ECCV} and GhostNet~\cite{Han_2020_CVPR} on the CIFAR-10 dataset, as shown in Fig~\ref{f2}.
The GhostSA bottleneck can reduce FLOPS by up to $50\times$, and GPU latency by up to $1.5\times$ while achieving similar or even higher accuracy. Among these backbones, GhostNet achieves the highest accuracy of $94.1\%$ within 18.1 milliseconds on the GPU, requiring more than 6.9 million FLOPs and 5.2 million parameters. After applying GhostSA in GhostNet, it only takes 13.5 ms to obtain an accuracy of $93.2\%$ on the GPU with 0.18M FLOPs and 1.6M parameters. We also compared the GhostSA module with ShiftAddNet~\cite{NEURIPS2020_1cf44d79}, and the results show that the GhostSA module is superior to the ShiftAddNet method.

\begin{table}[!htb]
\begin{center}
\resizebox{10.2cm}{!} {
\begin{tabular}{c|c|c|c|c|} 
\cline{2-5}
\multicolumn{1}{l|}{} & Model                                                             & Weights(M)                                                                         & FLOPs(M)                                                                           & Top-1 Acc. (\%)                                 \\ 
\cline{2-5}\noalign{\vskip\doublerulesep
         \vskip-\arrayrulewidth}\cline{2-5}
                      & \begin{tabular}[c]{@{}c@{}}ShuffleNetV2 0.5x~\cite{Zhang_2018_CVPR}\\\end{tabular} & \begin{tabular}[c]{@{}c@{}}1.4\\\end{tabular}                                      & \begin{tabular}[c]{@{}c@{}}42\\\end{tabular}                                       & \begin{tabular}[c]{@{}c@{}}61.1\\\end{tabular}  \\ 
% \cline{2-5}
                      & \begin{tabular}[c]{@{}c@{}}MobileNetV2 0.35x~\cite{Sandler_2018_CVPR} \\\end{tabular}      & \begin{tabular}[c]{@{}c@{}}1.7\\\end{tabular}                                      & \begin{tabular}[c]{@{}c@{}}59\\\end{tabular}                                       & \begin{tabular}[c]{@{}c@{}}60.3\\\end{tabular}  \\ 
% \cline{2-5}
                      & MobileNetV3 Small 0.75x~\cite{Howard_2019_ICCV}                                           & \begin{tabular}[c]{@{}c@{}}2.4\\\end{tabular}                                      & \begin{tabular}[c]{@{}c@{}}44\\\end{tabular}                                       & \begin{tabular}[c]{@{}c@{}}64.5\\\end{tabular}  \\ 
% \cline{2-5}
                      & \begin{tabular}[c]{@{}c@{}}GhostNet 0.5x~\cite{Han_2020_CVPR} \\\end{tabular}          & \begin{tabular}[c]{@{}c@{}}2.6\\\end{tabular}                                      & \begin{tabular}[c]{@{}c@{}}42\\\end{tabular}                                       & \begin{tabular}[c]{@{}c@{}}66.2\\\end{tabular}  \\ 
% \cline{2-5}
                      & \textcolor{blue}{GhostSANet 0.5x}                                 & \textcolor{blue}{2.3}                                                              & \textcolor{blue}{20}                                                               & \textcolor{blue}{66.9}                          \\ 
% \cline{2-5}
                                                                                                                                                                                                                                                                                        
                      & MobileNetV2 0.65X~\cite{Sandler_2018_CVPR}                                                 & \begin{tabular}[c]{@{}c@{}}2.3\\\end{tabular}                                      & \begin{tabular}[c]{@{}c@{}}172\\\end{tabular}                                      & \begin{tabular}[c]{@{}c@{}}67.2\\\end{tabular}  \\ 
% \cline{2-5}
                      & \begin{tabular}[c]{@{}c@{}}ShuffleNetV2 1.0X~\cite{Zhang_2018_CVPR}\\\end{tabular}       & \begin{tabular}[c]{@{}c@{}}2.3\\\end{tabular}                                      & \begin{tabular}[c]{@{}c@{}}146\\\end{tabular}                                      & \begin{tabular}[c]{@{}c@{}}67.8\\\end{tabular}  \\ 
% \cline{2-5}
                      & \begin{tabular}[c]{@{}c@{}}MobileNetV3 Large 0.75X~\cite{Howard_2019_ICCV}\\\end{tabular} & \begin{tabular}[c]{@{}c@{}}\textcolor{blue}{\textcolor{black}{4.0}}\\\end{tabular} & \begin{tabular}[c]{@{}c@{}}\textcolor{blue}{\textcolor{black}{155}}\\\end{tabular} & \begin{tabular}[c]{@{}c@{}}73.3\\\end{tabular}  \\ 
% \cline{2-5}
                      & \begin{tabular}[c]{@{}c@{}}GhostNet 1.0X~\cite{Han_2020_CVPR}\\\end{tabular}           & \begin{tabular}[c]{@{}c@{}}5.2\\\end{tabular}                                      & \begin{tabular}[c]{@{}c@{}}141\\\end{tabular}                                      & \begin{tabular}[c]{@{}c@{}}73.9\\\end{tabular}  \\ 
% \cline{2-5}
                      & \textcolor{blue}{GhostSANet~ 1.0X}                                & \textcolor{blue}{3.0}                                                              & \textcolor{blue}{40}                                                               & \textcolor{blue}{73.4}                          \\ 
% \cline{2-5}
                     
                      & \begin{tabular}[c]{@{}c@{}}MobileNetV2 1.0x~\cite{Sandler_2018_CVPR}\\\end{tabular}        & \begin{tabular}[c]{@{}c@{}}3.4\\\end{tabular}                                      & \begin{tabular}[c]{@{}c@{}}300\\\end{tabular}                                      & \begin{tabular}[c]{@{}c@{}}71.8\\\end{tabular}  \\ 
% \cline{2-5}
                      & \begin{tabular}[c]{@{}c@{}}ShuffleNetV2 1.5x~\cite{Zhang_2018_CVPR}\\\end{tabular}       & \begin{tabular}[c]{@{}c@{}}3.5\\\end{tabular}                                      & \begin{tabular}[c]{@{}c@{}}299\\\end{tabular}                                      & \begin{tabular}[c]{@{}c@{}}72.6\\\end{tabular}  \\ 
% \cline{2-5}
                      & FBNet-B~\cite{Wan_2020_CVPR}                                                           & 4.5                                                                                & 295                                                                                & 74.1                                            \\ 
% \cline{2-5}
                      & MnasNet-A1~\cite{Tan_2019_CVPR}                                                        & 3.9                                                                                & 312                                                                                & 75.2                                            \\ 
% \cline{2-5}
                      & MobileNetV3 Large 1X~\cite{Howard_2019_ICCV}                                              & \begin{tabular}[c]{@{}c@{}}5.4\\\end{tabular}                                      & \begin{tabular}[c]{@{}c@{}}219\\\end{tabular}                                      & \begin{tabular}[c]{@{}c@{}}75.2\\\end{tabular}  \\ 
% \cline{2-5}
                      & GhostNet 1.3x~\cite{Han_2020_CVPR}                                                     & \begin{tabular}[c]{@{}c@{}}7.3\\\end{tabular}                                      & \begin{tabular}[c]{@{}c@{}}226\\\end{tabular}                                      & \begin{tabular}[c]{@{}c@{}}75.7\\\end{tabular}  \\ 
% \cline{2-5}
                      & \textcolor{blue}{GhostSANet~ 1.3x}                                & \textcolor{blue}{3.7}                                                              & \textcolor{blue}{69}                                                               & \textcolor{blue}{76.4}                          \\
\cline{2-5}
\end{tabular}}
\end{center}
\caption{Comparison of state-of-the-art networks over classification accuracy, the number of weights and FLOPs on the ImageNet dataset.}
\label{imagenet}
\end{table}
\paragraph{\textbf{GhostSANet for ImageNet Classification}}
To evaluate the performance of the proposed GhostSANet model, we compared it to the SOTA models introduced in Fig~\ref{f2}, and add two other models into the comparison including FBNet V2~\cite{Wan_2020_CVPR} and MnasNet~\cite{Tan_2019_CVPR} on the ImageNet classification task. Following common practice, all networks have three levels of computational complexity, i.e., 40, 140, and 200-300 MFLOPs. For the sake of fairness, we test these models and GhostSANet under the same initial training settings. We set the DWS kernel size $k=3$ for the ghost part. The comparison results are summarised in Table~\ref{imagenet}, showing that greater FLOPs can lead to higher accuracy with a larger number of parameters. Our model provides improved performance compared to the SOTA, as shown in blue.
\paragraph{\textbf{Evaluation on Embedded Platform.}}
\begin{table}[!htb]
\begin{center}
\resizebox{10.2cm}{!} {
\begin{tabular}{|c|c|c|c|c|} 

\hline
Model name          & \begin{tabular}[c]{@{}c@{}}Backbone\\\textbf{G-Backbones}\end{tabular} & \begin{tabular}[c]{@{}c@{}}Top1-Acc.\\(\%)\end{tabular}                                                                             & \begin{tabular}[c]{@{}c@{}}GPU Speed \\(Batches/ms)\end{tabular}                  & \begin{tabular}[c]{@{}c@{}}CPU Speed\\(Images/sec)\end{tabular}                    \\ 
\hline\hline
                    & 0.25x(S)                                                        & \begin{tabular}[c]{@{}c@{}}\textcolor{blue}{\textcolor{black}{87.6}}\\\end{tabular} & \begin{tabular}[c]{@{}c@{}}\textcolor{blue}{\textcolor{black}{1207}}\\\end{tabular} & \begin{tabular}[c]{@{}c@{}}\textcolor{blue}{\textcolor{black}{114}}\\\end{tabular}  \\ 
% \cline{2-5}
                    & \textbf{G-0.25x(S)}                              & \textbf{88.3}                                                              & \textbf{1454}                                                              & \textbf{160}                                                               \\ 
% \cline{2-5}
MobileNet-V3~\cite{Howard_2019_ICCV}        & 0.75x(S)                                                        & \begin{tabular}[c]{@{}c@{}}91.3\\\end{tabular}                                      & \begin{tabular}[c]{@{}c@{}}1000\\\end{tabular}                                      & \begin{tabular}[c]{@{}c@{}}36\\\end{tabular}                                        \\ 
% \cline{2-5}
                    & \textbf{G-0.75x(S)}                              & \textbf{91.5}                                                              & \textbf{1084}                                                              & \textbf{74}                                                                \\ 
% \cline{2-5}
                    & 0.75x(Large)                                                        & \begin{tabular}[c]{@{}c@{}}91.9\\\end{tabular}                                      & \begin{tabular}[c]{@{}c@{}}888\\\end{tabular}                                       & \begin{tabular}[c]{@{}c@{}}24\\\end{tabular}                                        \\ 
% \cline{2-5}
                    & \textbf{G-0.75x(L)}                              & \textbf{91.9}                                                              & \textbf{1000}                                                              & \textbf{58}                                                                \\ 
\hline
                    & 0.25x                                                               & \begin{tabular}[c]{@{}c@{}}87.3\\\end{tabular}                                      & \begin{tabular}[c]{@{}c@{}}1049\\\end{tabular}                                      & \begin{tabular}[c]{@{}c@{}}81\\\end{tabular}                                        \\ 
% \cline{2-5}
                    & \textbf{G-0.25x}                                     & \textbf{87.8}                                                              & \textbf{1230}                                                              & \textbf{112}                                                               \\ 
% \cline{2-5}
MobileNet-V2~\cite{Sandler_2018_CVPR}        & 0.65x                                                               & \begin{tabular}[c]{@{}c@{}}89.5\\\end{tabular}                                      & \begin{tabular}[c]{@{}c@{}}429\\\end{tabular}                                       & \begin{tabular}[c]{@{}c@{}}36\\\end{tabular}                                        \\ 
% \cline{2-5}
                    & \textbf{G-0.65x}                                     & \textbf{90.3}                                                              & \textbf{500}                                                               & \textbf{38}                                                                \\ 
% \cline{2-5}
                    & 1x                                                                  & \begin{tabular}[c]{@{}c@{}}92.4\\\end{tabular}                                      & \begin{tabular}[c]{@{}c@{}}365\\\end{tabular}                                       & \begin{tabular}[c]{@{}c@{}}27\\\end{tabular}                                        \\ 
% \cline{2-5}
                    & \textbf{G-1x}                                        & \textbf{91.9}                                                              & \textbf{441}                                                               & \textbf{33}                                                                \\ 
\hline
                    & 0.5x                                                                & \begin{tabular}[c]{@{}c@{}}87.5\\\end{tabular}                                      & \begin{tabular}[c]{@{}c@{}}1032\\\end{tabular}                                      & \begin{tabular}[c]{@{}c@{}}96\\\end{tabular}                                        \\ 
% \cline{2-5}
ShuffleNet-V2~\cite{Zhang_2018_CVPR}       & \textbf{G-0.5x}                                      & \textbf{88.6}                                                              & \textbf{1306}                                                              & \textbf{152}                                                               \\ 
% \cline{2-5}
                    & 1x                                                                  & \begin{tabular}[c]{@{}c@{}}91.3\\\end{tabular}                                      & \begin{tabular}[c]{@{}c@{}}780\\\end{tabular}                                       & \begin{tabular}[c]{@{}c@{}}62\\\end{tabular}                                        \\ 
% \cline{2-5}
                    & \textbf{G-1x}                                        & \textbf{91.5}                                                              & \textbf{901}                                                               & \textbf{78}                                                                \\ 
\hline
GhostNet~\cite{Han_2020_CVPR}            & ~0.5x                                                               & \begin{tabular}[c]{@{}c@{}}91.6\\\end{tabular}                                      & \begin{tabular}[c]{@{}c@{}}727\\\end{tabular}                                       & \begin{tabular}[c]{@{}c@{}}28\\\end{tabular}                                        \\ 
% \cline{2-5}
                    & 1x                                                                  & \begin{tabular}[c]{@{}c@{}}94.8\\\end{tabular}                                      & \begin{tabular}[c]{@{}c@{}}627\\\end{tabular}                                       & \begin{tabular}[c]{@{}c@{}}19\\\end{tabular}                                        \\ 
\hline
\textcolor{blue}{GhostSANet} & \textcolor{blue}{0.5x}                                                       & \textcolor{blue}{92.1}                                                                       & \textcolor{blue}{1027}                                                                        & \textcolor{blue}{69}                                                                         \\ 
% \cline{2-5}
                    & \textcolor{blue}{1x}                                                         & \textcolor{blue}{95.1}                                                                       & \textcolor{blue}{819}                                                                        & \textcolor{blue}{56}                                                                         \\
\hline
\end{tabular}}
\end{center}
\caption{GhostSA applied backbone models on Jetson Nano for CIFAR10 that is shown as G-backbones. All benchmarks are implemented by us.}
\label{jetson}
\end{table}
In the final experiment, we test the GhostSA bottleneck and GhostSANet performance on the Jetson Nano. With limited computational resource, low latency requirements become essential for real-time data analysis. Therefore, our experiment focuses on evaluating the GPU latency (ms) and CPU latency (s) of the network. As summarized in Table~\ref{jetson}, we test the GhostSA bottleneck applied to each SOTA network, highlighted in bold. After applying the GhostSA bottleneck, the GPU/CPU speed of the backbone network can be increased by up to $1.3\times$ and $1.6\times$, respectively. Moreover, we test GhostSANet (shown in blue) showing that it can obtain a 0.3\% higher Top-1 test accuracy compared to GhostNet, with ~$1.3\times$ and ~$2\times$ lower GPU and CPU latency. In addition, GhostSANet requires less time to achieve similar performance, compared to other SOTA methods. For example, GhostSA-MobileNet V3 0.25x has lower GPU and CPU latency (44ms and 0.2s respectively), but the accuracy is only 88.3\%, while GhostSANet with the same accuracy only requires 26ms and 0.08s. Overall, our network is more effective and efficient on embedded devices than other current SOTA networks.
\section{Conclusions}\label{s5}
This paper proposes a new GhostSA module for building an efficient neural architecture, and shows that it can significantly concurrently reduce the computational cost, number of parameters, and inference latency. Moreover, the GhostSA module can flexibly adjust the number of shift and addition operations in a convolution layer through the hyper-parameter $\gamma$, which allows it to adapt to different hardware platforms. Experiments show that the proposed GhostSANet has excellent efficiency and accuracy on both desktop GPU/CPU and embedded platforms (Nvidia Jetson Nano). We anticipate that this work can inspire future work to design network architectures for embedded hardware systems that are both energy-saving and platform-aware.

\section{Acknowledgements}
This work was supported in part by the Engineering and Physical Sciences Research Council (EPSRC) under Grant EP/S030069/1.
\bibliography{egbib}

\begin{thebibliography}{10}

\bibitem{Zhang_2018_CVPR}
Xiangyu Zhang, Xinyu Zhou, Mengxiao Lin, and Jian Sun.
\newblock Shufflenet: An extremely efficient convolutional neural network for
  mobile devices.
\newblock In {\em Proceedings of the IEEE Conference on Computer Vision and
  Pattern Recognition (CVPR)}, June 2018.

\bibitem{Ma_2018_ECCV}
Ningning Ma, Xiangyu Zhang, Hai-Tao Zheng, and Jian Sun.
\newblock Shufflenet v2: Practical guidelines for efficient cnn architecture
  design.
\newblock In {\em Proceedings of the European Conference on Computer Vision
  (ECCV)}, September 2018.

\bibitem{howard2017mobilenets}
Andrew~G. Howard, Menglong Zhu, Bo~Chen, Dmitry Kalenichenko, Weijun Wang,
  Tobias Weyand, Marco Andreetto, and Hartwig Adam.
\newblock Mobilenets: Efficient convolutional neural networks for mobile vision
  applications.
\newblock {\em CoRR}, abs/1704.04861, 2017.

\bibitem{Sandler_2018_CVPR}
Mark Sandler, Andrew Howard, Menglong Zhu, Andrey Zhmoginov, and Liang-Chieh
  Chen.
\newblock Mobilenetv2: Inverted residuals and linear bottlenecks.
\newblock In {\em Proceedings of the IEEE Conference on Computer Vision and
  Pattern Recognition (CVPR)}, June 2018.

\bibitem{Howard_2019_ICCV}
Andrew Howard, Mark Sandler, Grace Chu, Liang-Chieh Chen, Bo~Chen, Mingxing
  Tan, Weijun Wang, Yukun Zhu, Ruoming Pang, Vijay Vasudevan, Quoc~V. Le, and
  Hartwig Adam.
\newblock Searching for mobilenetv3.
\newblock In {\em Proceedings of the IEEE/CVF International Conference on
  Computer Vision (ICCV)}, October 2019.

\bibitem{Wu2018ShiftAZ}
B.~Wu, Alvin Wan, Xiangyu Yue, Peter~H. Jin, S.~Zhao, Noah Golmant,
  A.~Gholaminejad, Joseph~E. Gonzalez, and K.~Keutzer.
\newblock Shift: A zero flop, zero parameter alternative to spatial
  convolutions.
\newblock {\em 2018 IEEE/CVF Conference on Computer Vision and Pattern
  Recognition (CVPR)}, pages 9127--9135, 2018.

\bibitem{Chen_2020_CVPR}
Hanting Chen, Yunhe Wang, Chunjing Xu, Boxin Shi, Chao Xu, Qi~Tian, and Chang
  Xu.
\newblock Addernet: Do we really need multiplications in deep learning?
\newblock In {\em Proceedings of the IEEE/CVF Conference on Computer Vision and
  Pattern Recognition (CVPR)}, June 2020.

\bibitem{Chen2019AllYN}
Weijie Chen, Di~Xie, Y.~Zhang, and Shiliang Pu.
\newblock All you need is a few shifts: Designing efficient convolutional
  neural networks for image classification.
\newblock {\em In IEEE/CVF Conference on Computer Vision and Pattern
  Recognition (CVPR)}, pages 7234--7243, 2019.

\bibitem{DBLP:conf/wacv/HeLZM19}
Yihui He, Xianggen Liu, Huasong Zhong, and Yuchun Ma.
\newblock Addressnet: Shift-based primitives for efficient convolutional neural
  networks.
\newblock In {\em {IEEE} Winter Conference on Applications of Computer Vision,
  {WACV} 2019, Waikoloa Village, HI, USA, January 7-11, 2019}, pages
  1213--1222. {IEEE}, 2019.

\bibitem{NEURIPS2020_1cf44d79}
Haoran You, Xiaohan Chen, Yongan Zhang, Chaojian Li, Sicheng Li, Zihao Liu,
  Zhangyang Wang, and Yingyan Lin.
\newblock Shiftaddnet: A hardware-inspired deep network.
\newblock In H.~Larochelle, M.~Ranzato, R.~Hadsell, M.~F. Balcan, and H.~Lin,
  editors, {\em Advances in Neural Information Processing Systems (NeurIPS)},
  volume~33, pages 2771--2783, 2020.

\bibitem{8766229}
IEEE.
\newblock Ieee standard for floating-point arithmetic.
\newblock {\em IEEE Std 754-2019 (Revision of IEEE 754-2008)}, pages 1--84,
  2019.

\bibitem{Han_2020_CVPR}
Kai Han, Yunhe Wang, Qi~Tian, Jianyuan Guo, Chunjing Xu, and Chang Xu.
\newblock Ghostnet: More features from cheap operations.
\newblock In {\em Proceedings of the IEEE/CVF Conference on Computer Vision and
  Pattern Recognition (CVPR)}, June 2020.

\bibitem{Elhoushi2019DeepShiftTM}
Mostafa Elhoushi, Zihao Chen, Farhan Shafiq, Ye~Henry Tian, and Joey~Yiwei Li.
\newblock Deepshift: Towards multiplication-less neural networks.
\newblock In {\em 2021 IEEE/CVF Conference on Computer Vision and Pattern
  Recognition Workshops (CVPRW)}, pages 2359--2368, 2021.

\bibitem{Song2020AdderSRTE}
Dehua Song, Yunhe Wang, Hanting Chen, Chang Xu, Chunjing Xu, and Dacheng Tao.
\newblock Addersr: Towards energy efficient image super-resolution.
\newblock In {\em Proceedings of the IEEE/CVF Conference on Computer Vision and
  Pattern Recognition (CVPR)}, pages 15648--15657, June 2021.

\bibitem{ReddiKK18}
Sashank~J. Reddi, Satyen Kale, and Sanjiv Kumar.
\newblock On the convergence of adam and beyond.
\newblock In {\em 6th International Conference on Learning Representations,
  {ICLR} 2018, Vancouver, BC, Canada, April 30 - May 3, 2018, Conference Track
  Proceedings}, 2018.

\bibitem{Jiabi}
Jia Bi and Steve~R. Gunn.
\newblock A variance controlled stochastic method with biased estimation for
  faster non-convex optimization.
\newblock In {\em European Conference on Machine Learning and Principles and
  Practice of Knowledge Discovery in Databases (ECML PKDD 2021)}, Sep 2021.

\bibitem{pmlr-v37-ioffe15}
Sergey Ioffe and Christian Szegedy.
\newblock Batch normalization: Accelerating deep network training by reducing
  internal covariate shift.
\newblock In Francis Bach and David Blei, editors, {\em Proceedings of the 32nd
  International Conference on Machine Learning}, volume~37 of {\em Proceedings
  of Machine Learning Research}, pages 448--456, Lille, France, 07--09 Jul
  2015. PMLR.

\bibitem{Krizhevsky09}
A.~Krizhevsky and G.~Hinton.
\newblock Learning multiple layers of features from tiny images.
\newblock {\em Master's thesis, Department of Computer Science, University of
  Toronto}, 2009.

\bibitem{ILSVRC15}
Olga Russakovsky, Jia Deng, Hao Su, Jonathan Krause, Sanjeev Satheesh, Sean Ma,
  Zhiheng Huang, Andrej Karpathy, Aditya Khosla, Michael Bernstein,
  Alexander~C. Berg, and Li~Fei-Fei.
\newblock {ImageNet Large Scale Visual Recognition Challenge}.
\newblock {\em International Journal of Computer Vision (IJCV)},
  115(3):211--252, 2015.

\bibitem{Simonyan15}
Karen Simonyan and Andrew Zisserman.
\newblock Very deep convolutional networks for large-scale image recognition.
\newblock In {\em International Conference on Learning Representations(ICLR)},
  2015.

\bibitem{7780459}
Kaiming He, Xiangyu Zhang, Shaoqing Ren, and Jian Sun.
\newblock Deep residual learning for image recognition.
\newblock In {\em 2016 IEEE Conference on Computer Vision and Pattern
  Recognition (CVPR)}, pages 770--778, 2016.

\bibitem{Wan_2020_CVPR}
Alvin Wan, Xiaoliang Dai, Peizhao Zhang, Zijian He, Yuandong Tian, Saining Xie,
  Bichen Wu, Matthew Yu, Tao Xu, Kan Chen, Peter Vajda, and Joseph~E. Gonzalez.
\newblock Fbnetv2: Differentiable neural architecture search for spatial and
  channel dimensions.
\newblock In {\em Proceedings of the IEEE/CVF Conference on Computer Vision and
  Pattern Recognition (CVPR)}, June 2020.

\bibitem{Tan_2019_CVPR}
Mingxing Tan, Bo~Chen, Ruoming Pang, Vijay Vasudevan, Mark Sandler, Andrew
  Howard, and Quoc~V. Le.
\newblock Mnasnet: Platform-aware neural architecture search for mobile.
\newblock In {\em Proceedings of the IEEE/CVF Conference on Computer Vision and
  Pattern Recognition (CVPR)}, June 2019.

\end{thebibliography}
\end{document}